\documentclass[10pt,twocolumn,letterpaper]{article}

\usepackage{cvpr}
\usepackage{times}
\usepackage{epsfig}
\usepackage{graphicx}
\usepackage{amsmath}
\usepackage{amssymb}

\usepackage{subcaption}

\usepackage[breaklinks=true,bookmarks=false]{hyperref}

\cvprfinalcopy 

\def\cvprPaperID{1626} 

\ifcvprfinal\fi

\begin{document}

\title{Learning Residual Images for Face Attribute Manipulation}

\author{Wei Shen \qquad Rujie Liu \\
Fujitsu Research \& Development Center, Beijing, China.
\\
{\tt\small \{shenwei, rjliu\}@cn.fujitsu.com}
}

\maketitle
\thispagestyle{empty}

\begin{abstract}
Face attributes are interesting due to their detailed description of human faces. Unlike prior researches working on attribute prediction, we address an inverse and more challenging problem called face attribute manipulation which aims at modifying a face image according to a given attribute value. 
Instead of manipulating the whole image, we propose to learn the corresponding residual image defined as the difference between images before and after the manipulation. In this way, the manipulation can be operated efficiently with modest pixel modification. 
The framework of our approach is based on the Generative Adversarial Network. It consists of two image transformation networks and a discriminative network. The transformation networks are responsible for the attribute manipulation and its dual operation and the discriminative network is used to distinguish the generated images from real images. 
We also apply dual learning to allow transformation networks to learn from each other.
Experiments show that residual images can be effectively learned and used for attribute manipulations. The generated images remain most of the details in attribute-irrelevant areas. 
\end{abstract}

\section{Introduction}
\begin{figure}[t]
\begin{center}
\begin{minipage}[b]{0.48\textwidth}
\includegraphics[width=1\linewidth]{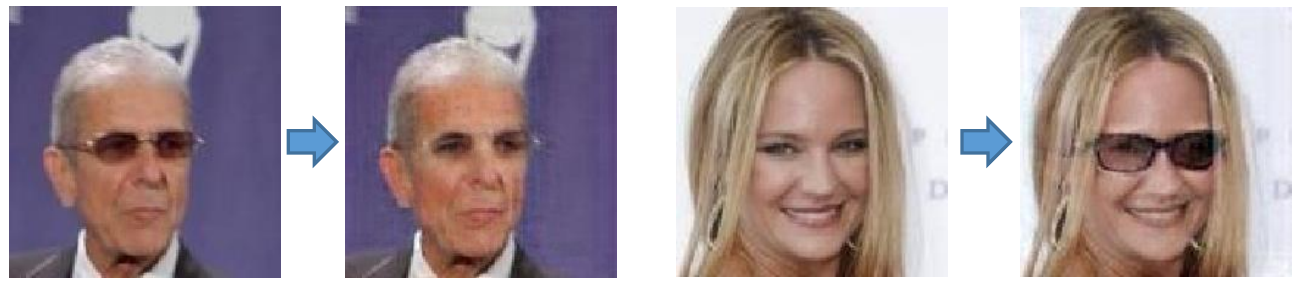}
\subcaption{\textit{Glasses}: remove and add the glasses}
\end{minipage}
\begin{minipage}[b]{0.48\textwidth}
\includegraphics[width=1\linewidth]{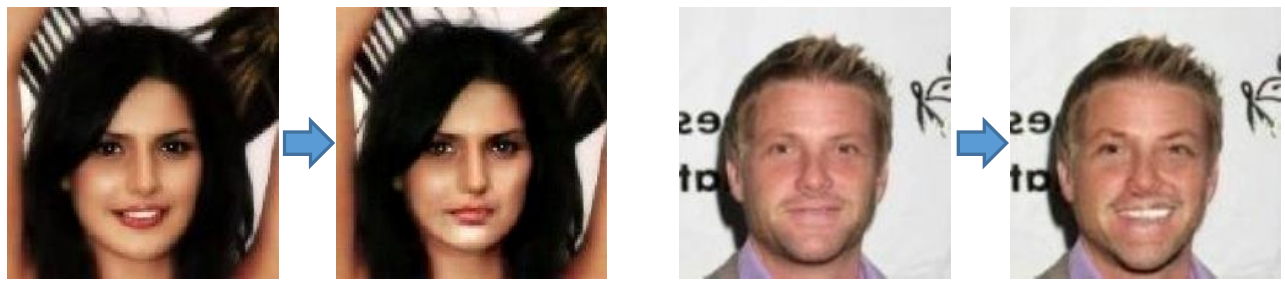}
\subcaption{\textit{Mouth\_open}: close and open the mouth}
\end{minipage}
\begin{minipage}[b]{0.48\textwidth}
\includegraphics[width=1\linewidth]{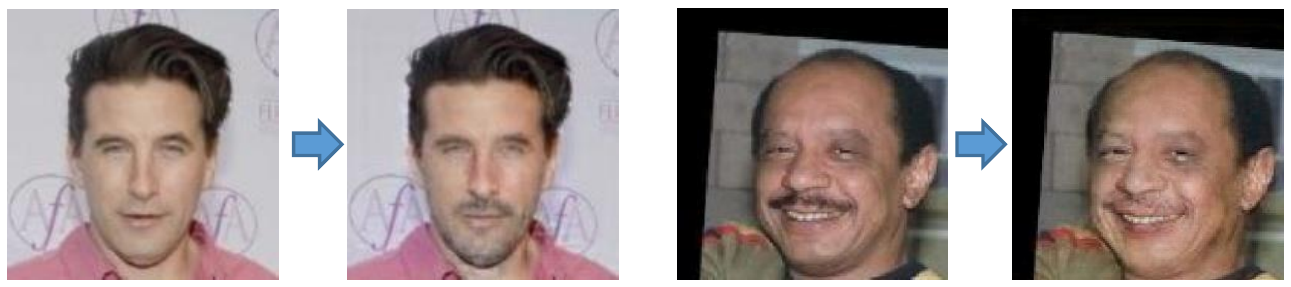}
\subcaption{\textit{No\_beard}: add and remove the beard}
\end{minipage}

\end{center}
   \caption{Illustration of face attribute manipulation. 
   From top to bottom are the manipulations of \textit{glasses}, \textit{mouth\_open} and \textit{no\_beard}.}
\label{fig:demo}
\end{figure}

Considerable progresses have been made on face image processing, such as age analysis~\cite{niu2016ordinal}\cite{Wang_2016_CVPR}, emotion detection~\cite{benitez2016emotionet}\cite{eleftheriadis2015discriminative} and attribute classification~\cite{Ehrlich_2016_CVPR_Workshops}\cite{liu2015deep}\cite{kumar2009attribute}\cite{li2016smile}. Most of these studies concentrate on inferring attributes from images. However, we raise an inverse question on whether we can manipulate a face image towards a desired attribute value (i.e. face attribute manipulation). Some examples are shown in Fig.~\ref{fig:demo}. 

Generative models such as generative adversarial networks (GANs)~\cite{goodfellow2014generative} and variational autoencoders (VAEs)~\cite{kingma2013auto} are powerful models capable of generating images. Images generated from GAN models are sharp and realistic. However, they can not encode images since it is the random noise that is used for image generation. Compared to GAN models, VAE models are able to encode the given image to a latent representation. Nevertheless, passing images through the encoder-decoder pipeline often harms the quality of the reconstruction. In the scenario of face attribute manipulation, those details can be identity-related and the loss of those details will cause undesired changes. Thus, it is difficult to directly employ GAN models or VAE models to face attribute manipulation.

An alternative way is to view face attribute manipulation as a transformation process which takes in original images as input and then outputs transformed images without explicit embedding. Such a transformation process can be efficiently implemented by a feed-forward convolutional neural network (CNN). 
When manipulating face attributes, the feed-forward network is required to modify the attribute-specific area and keep irrelevant areas unchanged, both of which are challenging. 

\begin{figure*}
\begin{center}
\includegraphics[width=1.0\linewidth]{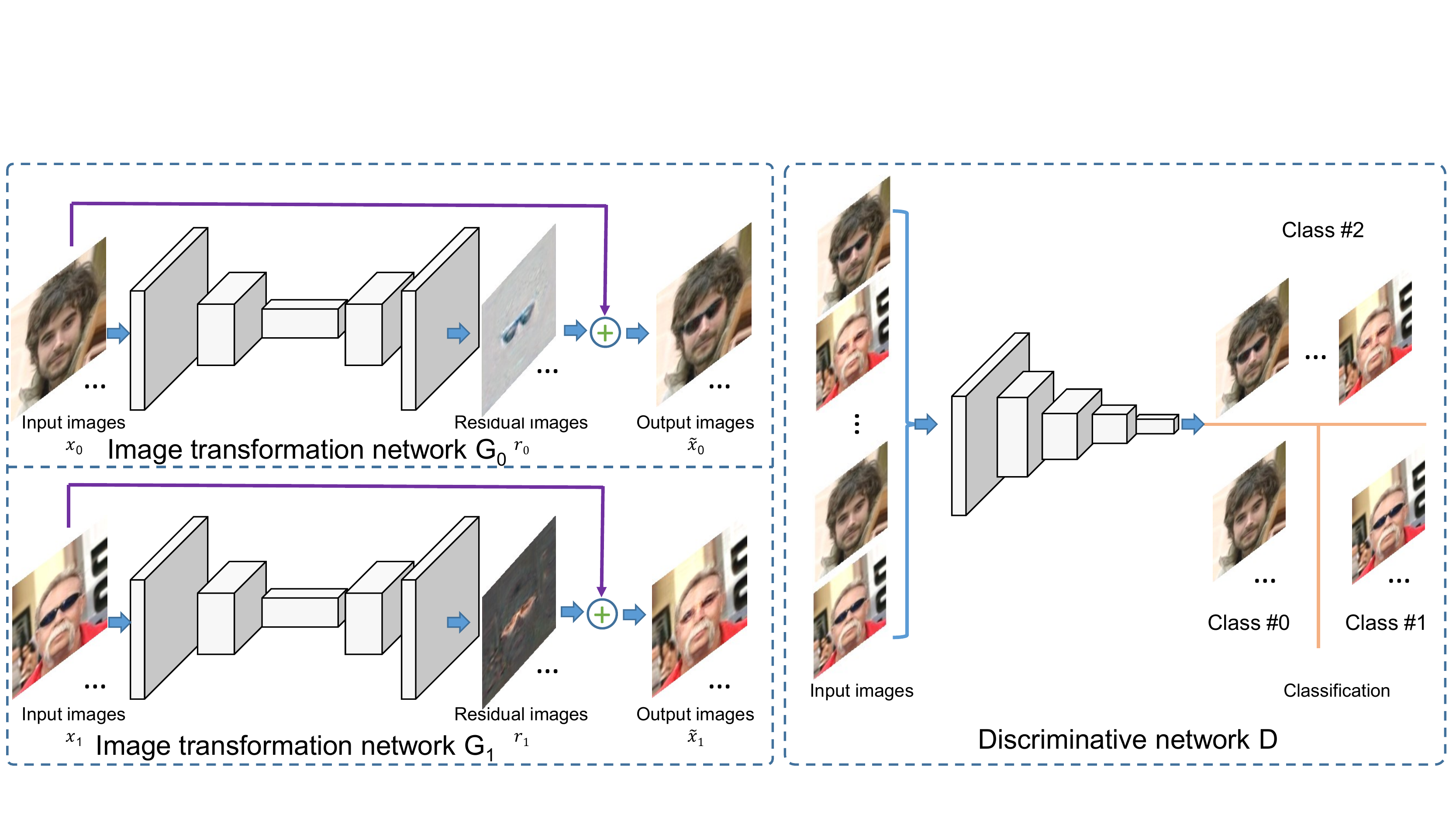}
\end{center}
   \caption{The architecture of the proposed method. Two image transformation networks $G_0$ and $G_1$ perform the inverse attribute manipulation (\ie adding glasses and removing glasses). Both $G_0$ and $G_1$ produce residual images with reference to the input images. The final output images are the pixel-wise addition of the residual images and the input images. The discriminative network $D$ is a three-category classifier that classifies images from different categories (\ie images generated from $G_0$ and $G_1$, images with positive attribute labels, and images with negative attribute labels).}
\label{fig:arch}
\end{figure*}

In this paper, we propose a novel method based on residual image learning for face attribute manipulation. The method combines the generative power of the GAN model with the efficiency of the feed-forward network (see Fig.~\ref{fig:arch}). 
We model the manipulation operation as learning the residual image which is defined as the difference between the original input image and the desired manipulated image. Compared to learning the whole manipulated image, learning only the residual image avoids the redundant attribute-irrelevant information by concentrating on the essential attribute-specific knowledge. To improve the efficiency of manipulation learning, we adopt two CNNs to model two inverse manipulations (\eg removing glasses as the primal manipulation and adding glasses as the dual manipulation, Fig.~\ref{fig:arch}) and apply the strategy of dual learning during the training phase. Our contribution can be summarized as follows.

\begin{enumerate}
\item We propose to learn residual images for face attribute manipulation. The proposed method focuses on the attribute-specific face area instead of the entire face which contains many redundant irrelevant details.


\item We devise a dual learning scheme to learn two inverse attribute manipulations (one as the primal manipulation and the other as the dual manipulation) simultaneously. We demonstrate that the dual learning process is helpful for generating high quality images.

\item Though it is difficult to assess the manipulated images quantitatively, we adopt the landmark detection accuracy gain as the metric to quantitatively show the effectiveness of the proposed method for glasses removal.

\end{enumerate}

\section{Related Work}

There are many techniques for image generation in recent years~\cite{radford2015unsupervised}\cite{chen2016infogan}\cite{larsen2015autoencoding}\cite{gregor2015draw}\cite{denton2015deep}\cite{kingma2013auto}. Radford \etal~\cite{radford2015unsupervised} applied deep convolutional generative adversarial networks (DCGANs) to learn a hierarchy of representations from object parts to scenes for general image generation. Chen \etal~\cite{chen2016infogan} introduced an information-theoretic extension to the GAN that was able to learn disentangled representations. Larsen \etal~\cite{larsen2015autoencoding} combined the VAE with the GAN to learn an embedding in which high-level abstract visual features could be modified using simple arithmetic.

Our work is an independent work along with~\cite{li2016deep}. In ~\cite{li2016deep}, Li \etal proposed a deep convolutional network model for identity-aware transfer of facial attributes. The differences between our work and ~\cite{li2016deep} are noticeable in three aspects. (1) Our method generates manipulated images using residual images which is different from ~\cite{li2016deep}. (2) Our method models two inverse manipulations within one single architecture by sharing the same discriminator while the work in~\cite{li2016deep} treats each manipulation independently. (3) Our method does not require post-processing which is essential in ~\cite{li2016deep}.

\section{Learning the Residual Image}
The architecture of the proposed method is presented in Fig.~\ref{fig:arch}. For each face attribute manipulation, it contains two image transformation networks $G_0$ and $G_1$ and a discriminative network $D$. $G_0$ and $G_1$ simulate the primal and the dual manipulation respectively. $D$ classifies the reference images and generated images into three categories. The following sections will first give a brief introduction of the generative adversarial network and then the detailed description of the proposed method.

\subsection{Generative Adversarial Networks}
The generative adversarial network is introduced by Goodfellow \etal~\cite{goodfellow2014generative}. It is an unsupervised framework containing a generative model $G$ and a discriminative model $D$. The two models play a minimax two-player game in which $G$ tries to recover the training data and fool $D$ to make a mistake about whether the data is from realistic data distribution or from $G$. Given a data prior $p_{data}$ on data $x$ and a prior on the input noise variables $p_z(z)$, the formal expression of the minimax game is as follows:

\begin{equation}
\begin{split}
\min \limits_{G}\max \limits_{D}V(D,G)=E_{x\sim p_{data}(x)}[\log D(\mathbf{x})]
\\
+E_{z\sim p_z(z)}[\log (1-D(G(z)))].
\end{split}
\end{equation}

The parameters of both $G$ and $D$ are updated iteratively during the training process. The GAN framework provides an effective way to learn data distribution of realistic images and makes it possible to generate images with desired attribute. Based on the GAN framework, we redesign the generator and the discriminator for face attribute manipulation in the following sections.

\subsection{Image Transformation Networks}
The motivation of our approach is that face attribute manipulation usually only needs modest modification of the attribute-specific face area while other parts remain unchanged. For example, when removing a pair of glasses from a face image, only the area of the glasses should be replaced with face skin or eyes while the change of other face parts such as mouth, nose, and hair should not be involved. Thus, we model the manipulation as learning the residual image targeted on the attribute-specific area. 

As shown in Fig. ~\ref{fig:arch}, image transformation networks $G_0$ and $G_1$ are used to simulate the manipulation and its dual operation. Given the input face image $x_0$ with a negative attribute value and the input face image $x_1$ with a positive attribute value, the learned network $G_0$ and $G_1$ apply the manipulation transformations to yield the residual images $r_0$ and $r_1$. Then the input images are added to the residual images as the final outputs $\tilde{x}_0$ and $\tilde{x}_1$:
\begin{equation}
\tilde{x}_i=x_i+r_i=x_i+G_i(x_i), i=0,1.
\end{equation}

In order to let the residual image be sparse, we apply an L-1 norm regularization as
\begin{equation}
\ell_{pix}(r_i)=||r_i||_1, i=0,1.
\end{equation}


\subsection{The Discriminative Network}
Given the real images $x_0$ and $x_1$ with known attribute label $0$ and label $1$, we regard the transformed images $\tilde{x}_0$ and $\tilde{x}_1$ as an extra category with label $2$. The loss function is:
\begin{equation}
\ell_{cls}(t, p)=-\log (p_t), t=0,1,2,
\end{equation}
where $t$ is the label of the image and $p_t$ is the softmax probability of the $t$-th label. Similar strategy for constructing the GAN loss is also adopted in ~\cite{salimans2016improved}.

Perceptual loss is widely used to measure the content difference between different  images~\cite{johnson2016perceptual}\cite{gatys2016image}\cite{larsen2015autoencoding}. We also apply this loss to encourage the transformed image to have similar content to the input face image. Let $\phi(x)$ be the activation of the third layer in $D$. The perceptual loss is defined as:
\begin{equation}
\ell_{per}(x, \tilde{x})=||\phi(x)-\phi(\tilde{x})||_1.
\end{equation}

Given the discriminative network $D$, the GAN loss for the image transformation networks $G_0$ and $G_1$ is
\begin{equation}
\ell_{GAN}=
\begin{cases}
-\log (D(G_i(x_i))) & i = 0,\\
-\log(1-D(G_i(x_i))) & i = 1.
\end{cases}
\end{equation}

\subsection{Dual Learning}
In addition to applying adversarial learning in the model training, we also adopt dual learning which has been successfully applied in machine translation~\cite{xia2016dual}. A brief introduction is as follows. Any machine translation has a dual task, \ie the source language to the target language (primal) and the target language to the source language (dual). The mechanism of dual learning can be viewed as a two-player communication game. The first player translates a message from language A to language B and sends it to the second player. The second player checks if it is natural in language B and notifies the first player. Then he translates the message to language A and sends it back to the first player. The first player checks whether the received message is consistent with his original message and notifies the second player. The information feedback signals from both players can benefit each other through a closed loop. 

\begin{figure}
\begin{center}
\includegraphics[width=1.0\linewidth]{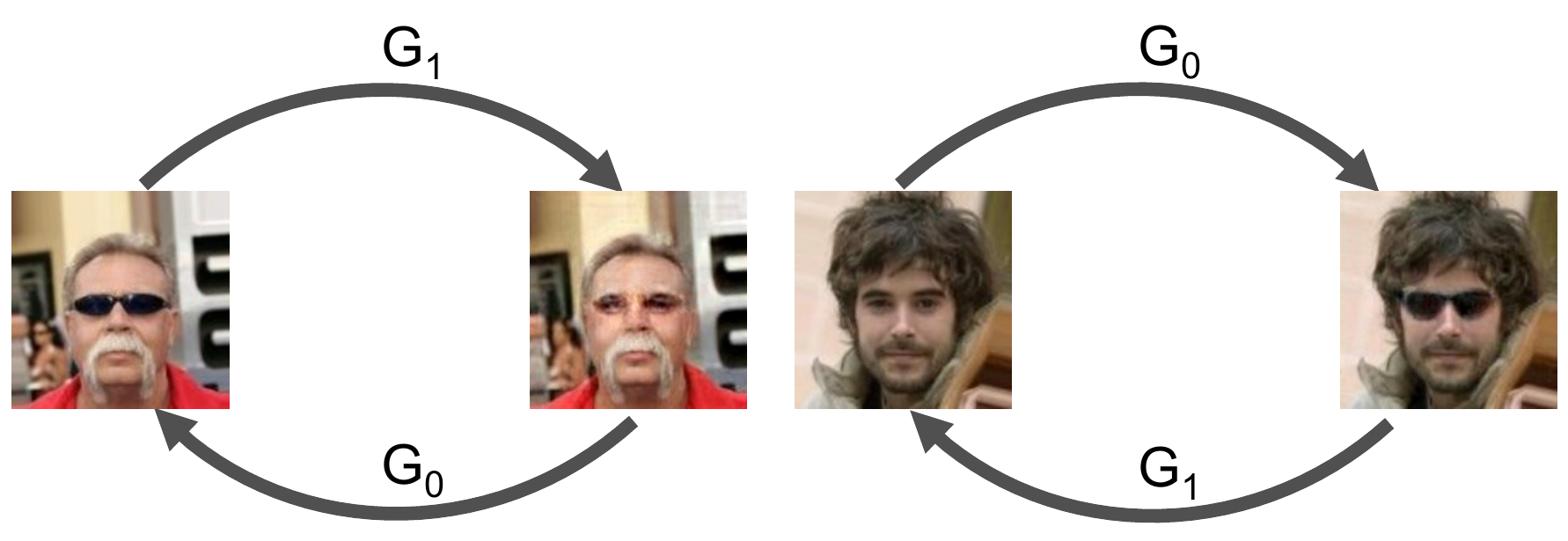}
\end{center}
   \caption{The dual learning process in this work.}
\label{fig:dual}
\end{figure}

The dual learning process in this work is implemented as Fig.~\ref{fig:dual}. For a given image $x_0$ with a negative attribute value, we pass it through the transformation network $G_0$. The obtained image $\tilde{x}_0=G_0(x_0)$ is then fed to the transformation network $G_1$. 
The yielded image is $\hat{x}_0=G_1(\tilde{x}_0)=G_1(G_0(x_0))$. 
Since $G_0$ and $G_1$ are the primal task and the dual task respectively, $\hat{x}_0$ is expected to have the same attribute value as $x_0$. Similar process is also applied for $x_1$. The loss function for the transformation networks in this phase is expressed as:
\begin{equation}
\ell_{dual}(\tilde{x}_i) = 
\begin{cases}
-\log (1-D(G_{1-i}(\tilde{x}_i))) & i=0,\\
-\log (D(G_{1-i}(\tilde{x}_i))) & i=1.
\end{cases}
\end{equation}

\subsection{Loss Function}
Taking the loss functions all together, we have the following loss function for $G_0$/$G_1$:
\begin{equation}
\ell_{G}=\ell_{GAN} + \ell_{dual} + \alpha \ell_{pix}+\beta \ell_{per},
\end{equation}
where $\alpha$ and $\beta$ are constant weight for regularization terms. 

For $D$, the loss function is
\begin{equation}
\ell_D=\ell_{cls}.
\end{equation}


\section{Datasets}
We adopt two datasets in our experiments, \ie the CelebA dataset~\cite{liu2015faceattributes} and the Labeled Faces in the Wild (LFW) dataset~\cite{LFWTech}. The CelebA dataset contains more than 200K celebrity images, each with 40 binary attributes. We pick 6 of them, \ie \textit{glasses}, \textit{mouth\_open}, \textit{smile}, \textit{no\_beard}, \textit{young}, and \textit{male} to evaluate the proposed method. The center part of the aligned images in the CelebA dataset are cropped and scaled to 128$\times$128. Despite there are a large number of images in the dataset, the attribute labels are highly biased. Thus, for each attribute, 1,000 images from the attribute-positive class and 1,000 images from the attribute-negative class are randomly selected for test. From the rest images, we select all the images belong to the minority class and equal number of images from the majority class to make a balance dataset. The LFW dataset is used only for testing the generalization of our method. Note that there are no ground truth manipulated images in the CelebA dataset for training the transformation networks. 

\section{Implementation Details}
\begin{table*}
\begin{center}
\begin{tabular}{|c|c|}
\hline
Image transformation networks $G_0$/$G_1$ & Discriminative network $D$\\
\hline
\hline
Input 128$\times$128 color images & Input 128$\times$128 color images \\
\hline
 5$\times$5 conv. 64 leaky RELU. stride 1. batchnorm 
& 4$\times$4 conv. 64 leaky RELU. stride 2. batchnorm\\
 \hline
 4$\times$4 conv. 128 leaky RELU. stride 2. batchnorm  
 & 4$\times$4 conv. 128 leaky RELU. stride 2. batchnorm\\
 \hline
 4$\times$4 conv. 256 leaky RELU. stride 2. batchnorm 
 & 4$\times$4 conv. 256 leaky RELU. stride 2. batchnorm\\
 \hline
 3$\times$3 conv. 128 leaky RELU. stride 1. upsampling. batchnorm
 & 4$\times$4 conv. 512 leaky RELU. stride 2. batchnorm\\
 \hline
 3$\times$3 conv. 64 leaky RELU. stride 1. upsampling. batchnorm
 & 4$\times$4 conv. 1024 leaky RELU. stride 2. batchnorm\\
 \hline
 4$\times$4 conv. 3 
 & 4$\times$4 conv. 1\\
\hline
\end{tabular}
\end{center}
\caption{The network architectures of the image transformation networks $G_0$/$G_1$ and the discriminative network $D$}
\label{tb:detail_arch}
\end{table*}

The detailed architectures of $G_0$, $G_1$ and $D$ are specified in Tab.~\ref{tb:detail_arch}. We keep $\beta = 0.1\alpha$ and set $\alpha$=5e-4 for local face attribute (\ie \textit{glasses, no\_beard, mouth\_open, smile}) manipulation and $\alpha$=1e-6 for global face attribute (\ie \textit{male, young}) manipulation.  The weights of all the networks are initialized from a zero-centered Normal distribution with standard deviation 0.02. The Adam optimizer~\cite{kingma2014adam} is used in the training phase. The learning rates for both the transformation networks and the discriminators are the same 2e-4. Both $G_0$ and $G_1$ are trained at the same time without any staging.

\begin{figure*}
\begin{center}
\begin{minipage}[b]{0.49\textwidth}  
\includegraphics[width=1\linewidth]{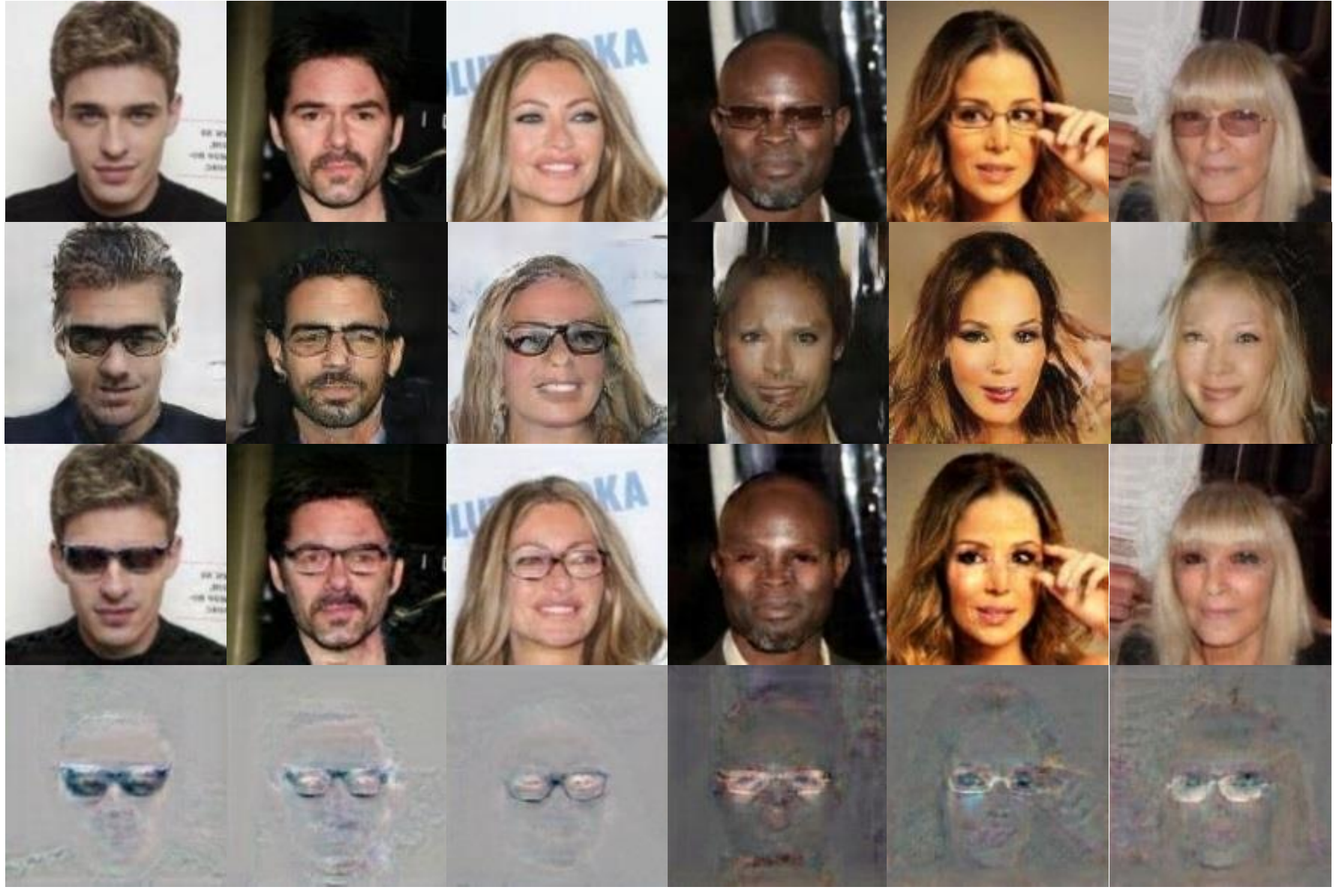}
\subcaption{\textit{Glasses}}
\end{minipage}
\begin{minipage}[b]{0.49\textwidth}
\includegraphics[width=1\linewidth]{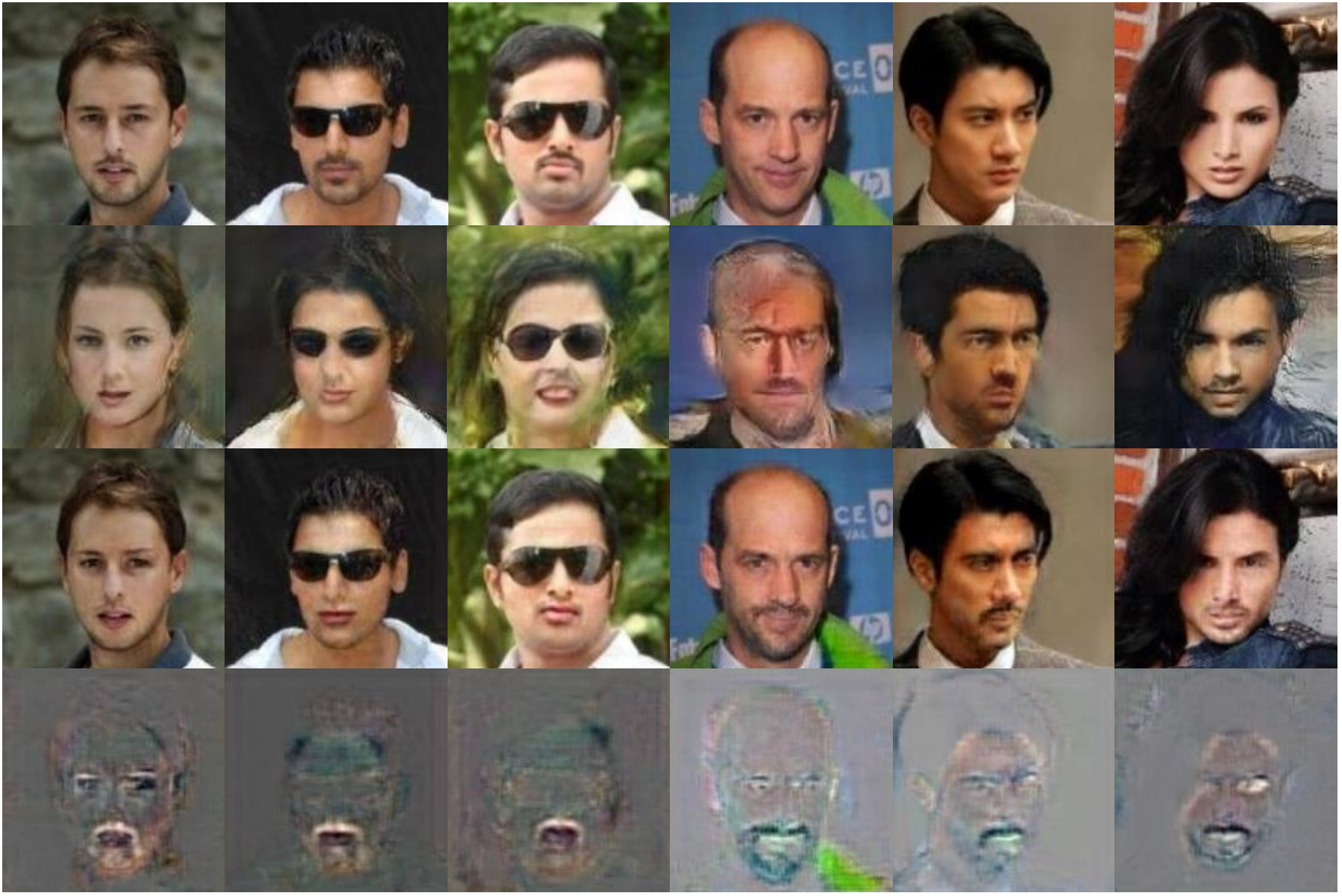}
\subcaption{\textit{No\_beard}}
\end{minipage}
\begin{minipage}[b]{0.49\textwidth}
\includegraphics[width=1\linewidth]{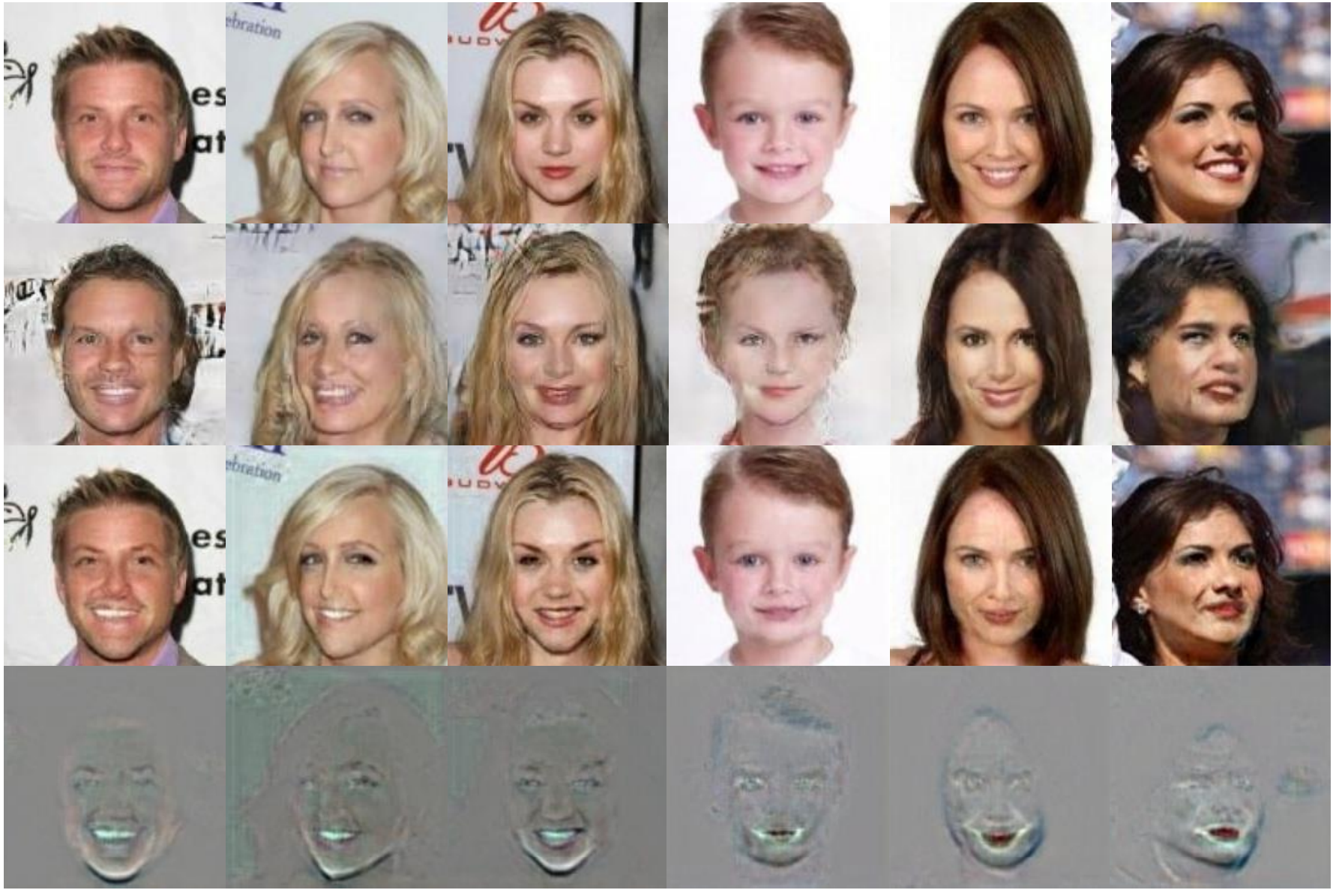}
\subcaption{\textit{Mouth\_open}}
\end{minipage}
\begin{minipage}[b]{0.49\textwidth}
\includegraphics[width=1\linewidth]{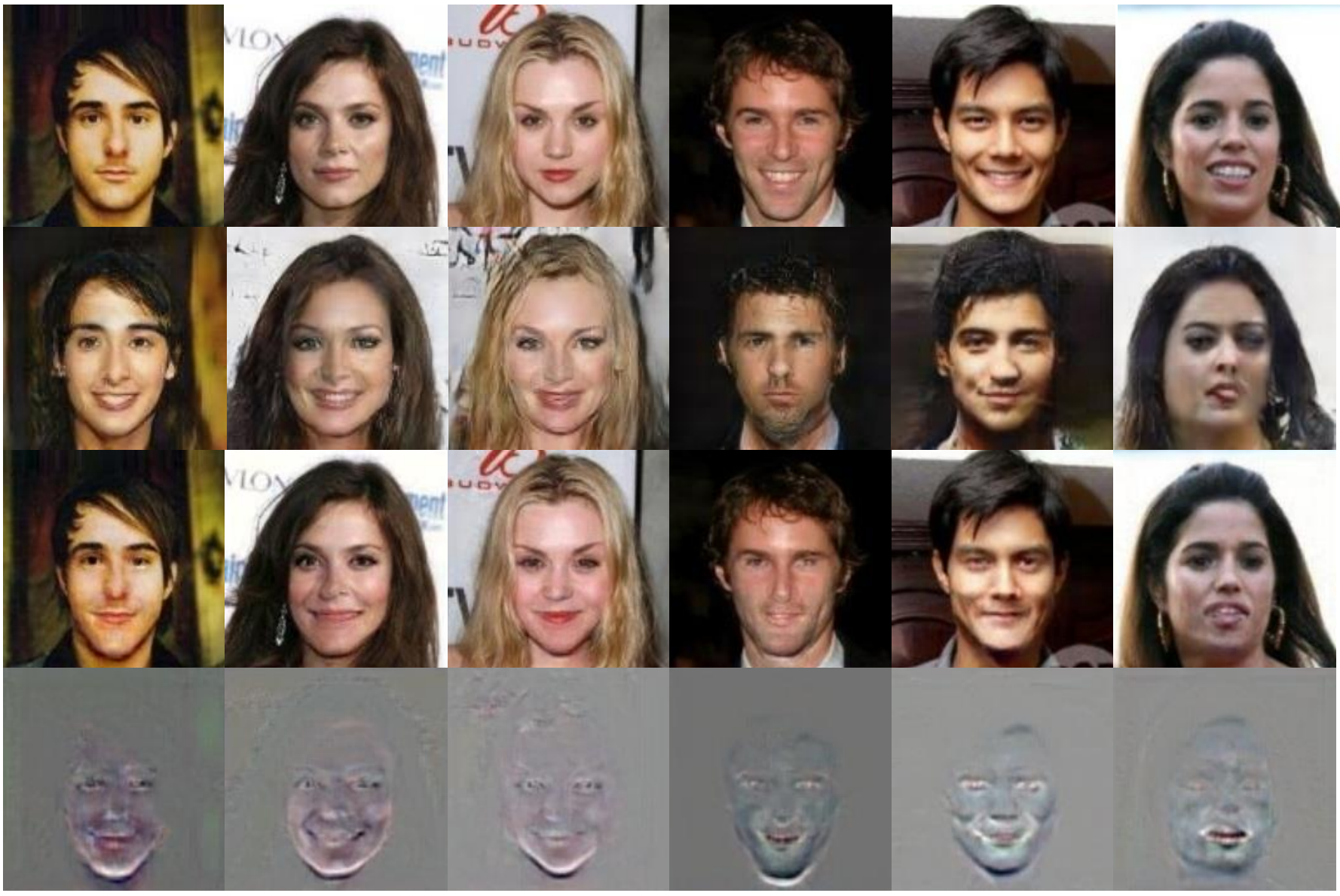}
\subcaption{\textit{Smile}}
\end{minipage}
\begin{minipage}[b]{0.49\textwidth}
\includegraphics[width=1\linewidth]{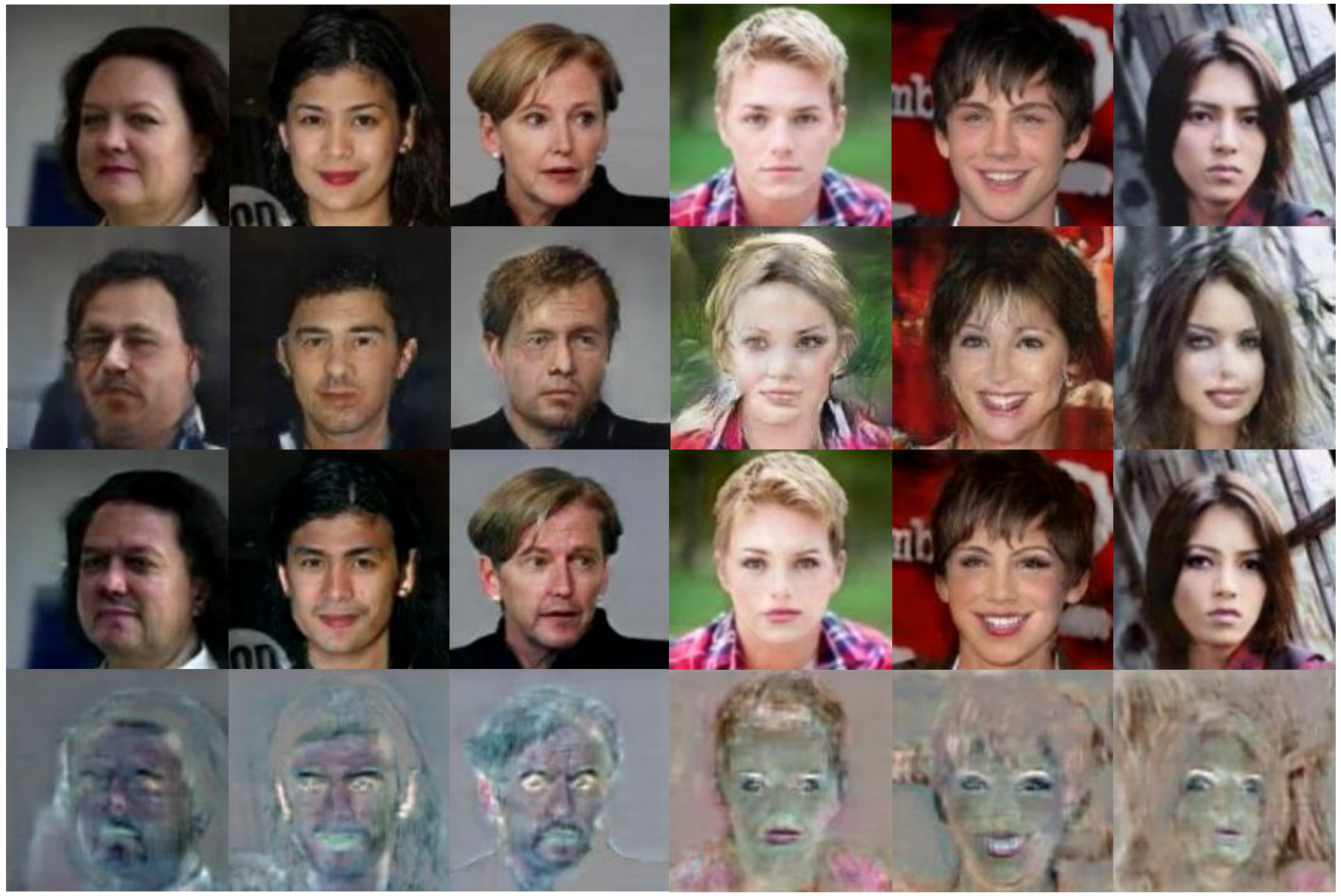}
\subcaption{\textit{Male}}
\end{minipage}
\begin{minipage}[b]{0.49\textwidth}
\includegraphics[width=1\linewidth]{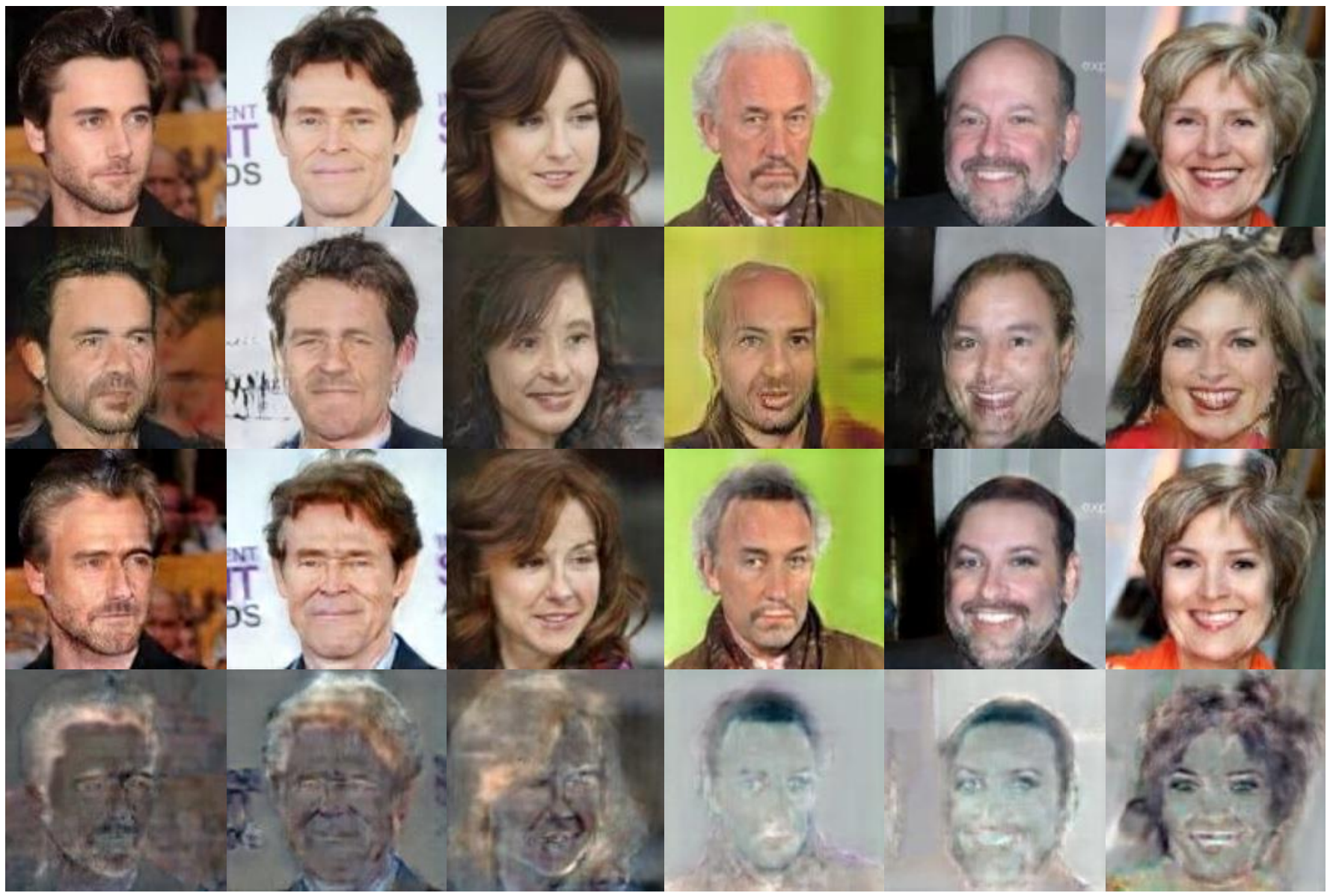}
\subcaption{\textit{Young}}
\end{minipage}
\end{center}
   \caption{Face attribute manipulation on the CelebA dataset. For each sub-figure, the first row shows the original face images. The second and the third row are the manipulated images using the VAE-GAN model~\cite{larsen2015autoencoding} and the proposed method respectively. The last row illustrates the residual images learned by the proposed method. Results of the primal/dual task of each attribute manipulation are presented in the first/last three columns. }
\label{fig:res}
\end{figure*}

\begin{figure*}
\begin{center}
\begin{minipage}[b]{0.49\textwidth}  
\includegraphics[width=1\linewidth]{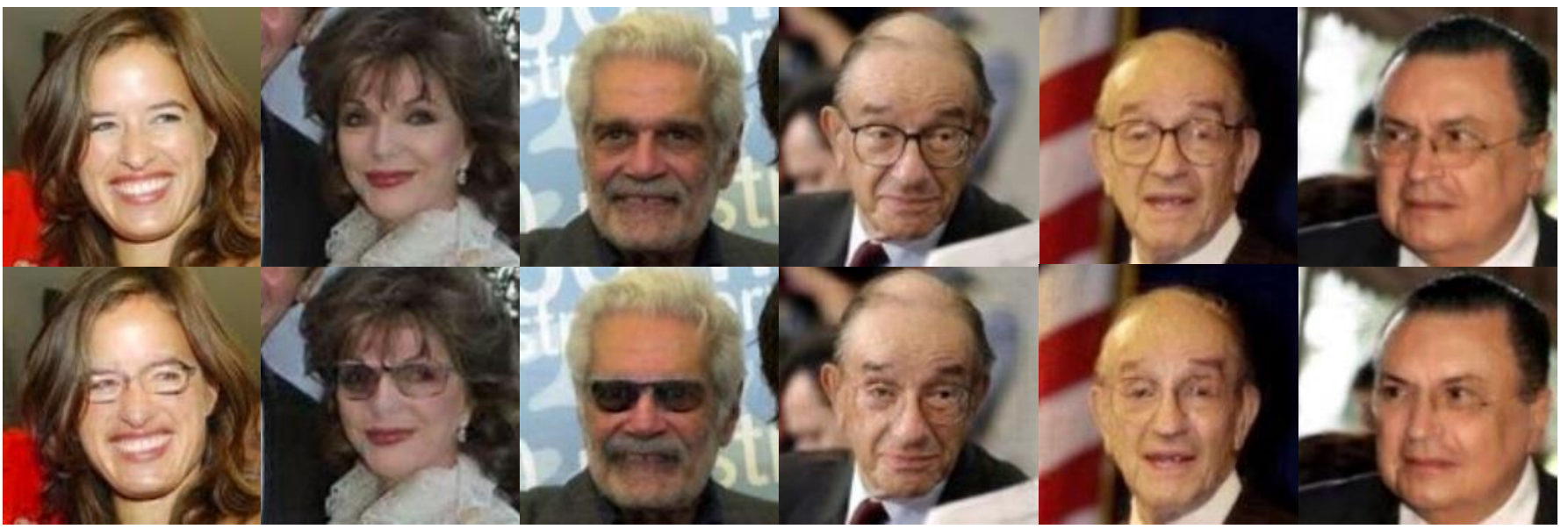}
\subcaption{\textit{Glasses}}
\end{minipage}
\begin{minipage}[b]{0.49\textwidth}
\includegraphics[width=1\linewidth]{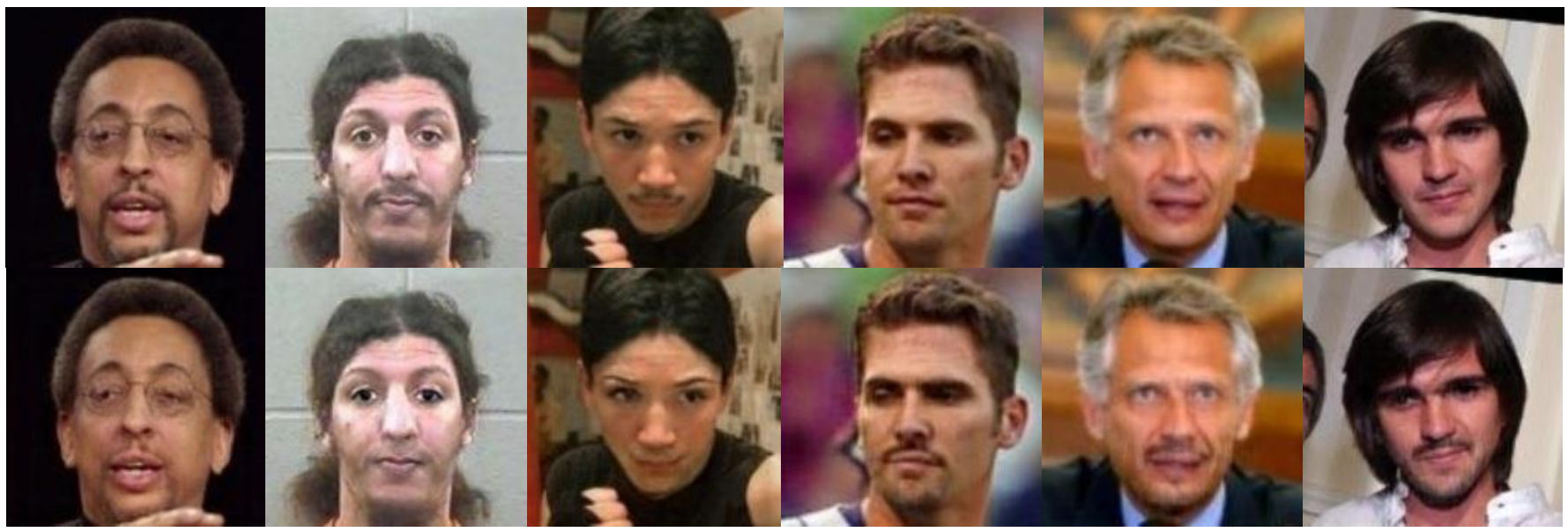}
\subcaption{\textit{No\_beard}}
\end{minipage}
\begin{minipage}[b]{0.49\textwidth}
\includegraphics[width=1\linewidth]{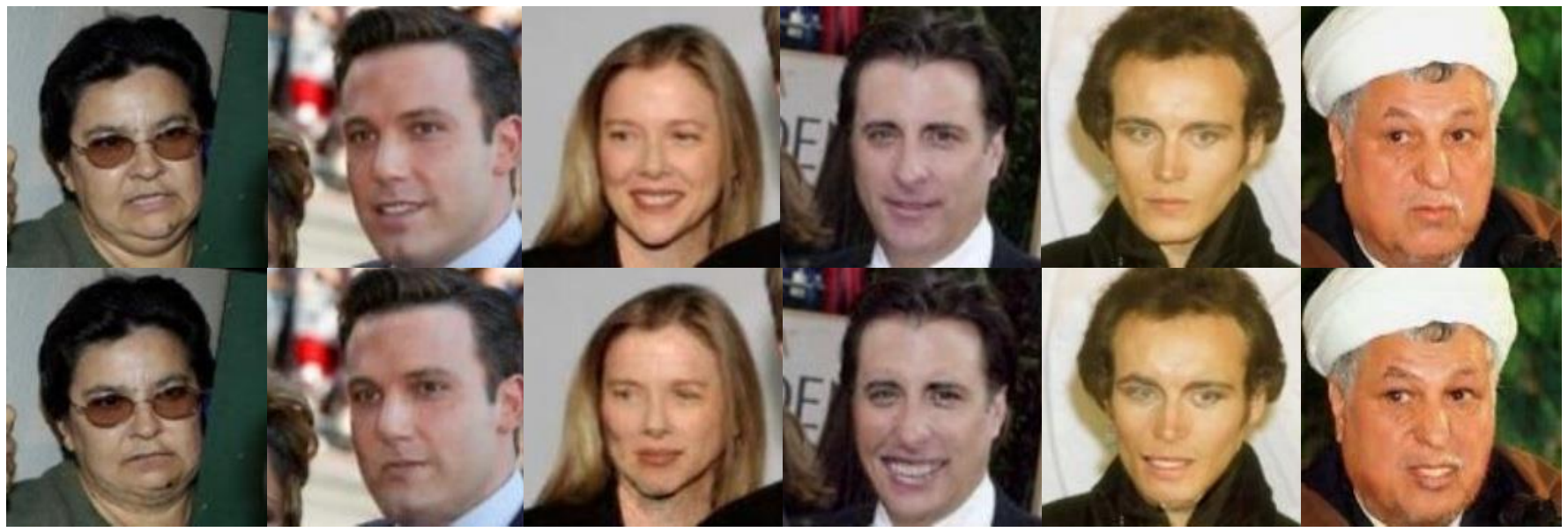}
\subcaption{\textit{Mouth\_open}}
\end{minipage}
\begin{minipage}[b]{0.49\textwidth}
\includegraphics[width=1\linewidth]{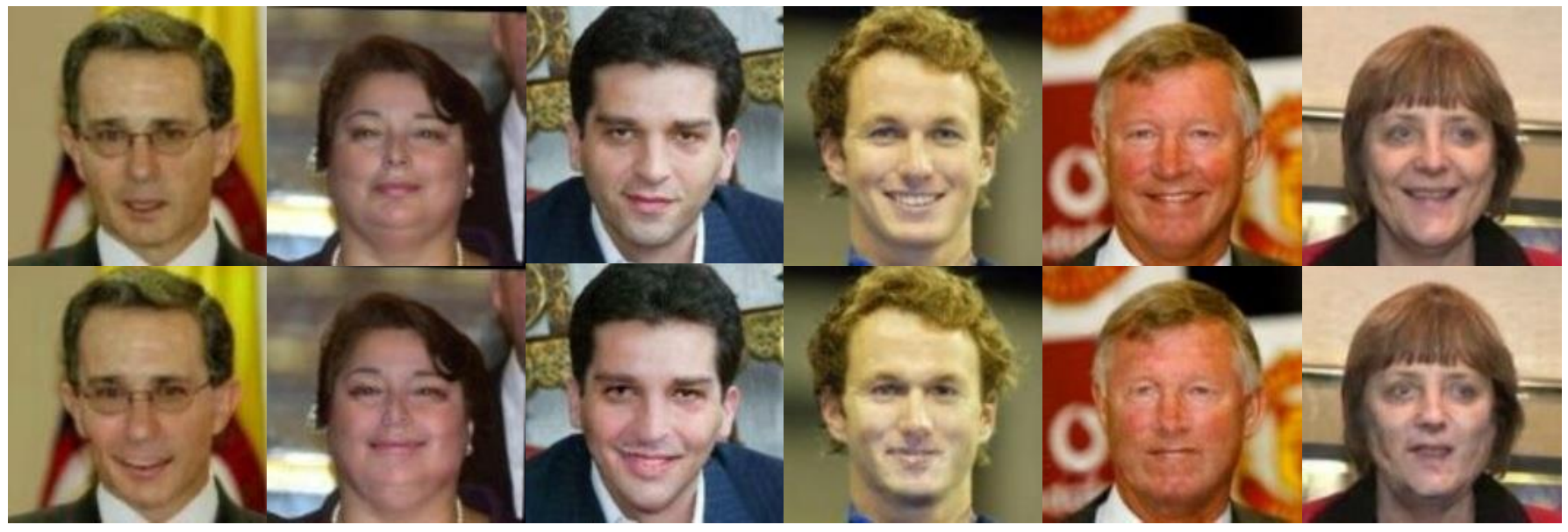}
\subcaption{\textit{Smile}}
\end{minipage}
\begin{minipage}[b]{0.49\textwidth}
\includegraphics[width=1\linewidth]{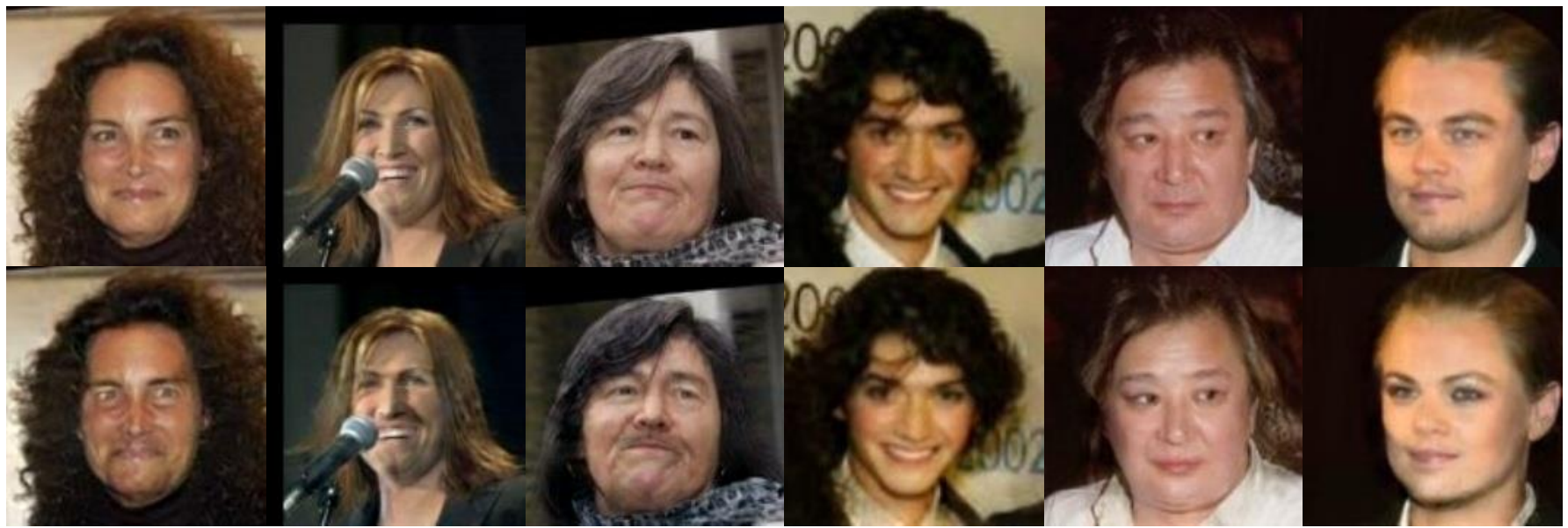}
\subcaption{\textit{Male}}
\end{minipage}
\begin{minipage}[b]{0.49\textwidth}
\includegraphics[width=1\linewidth]{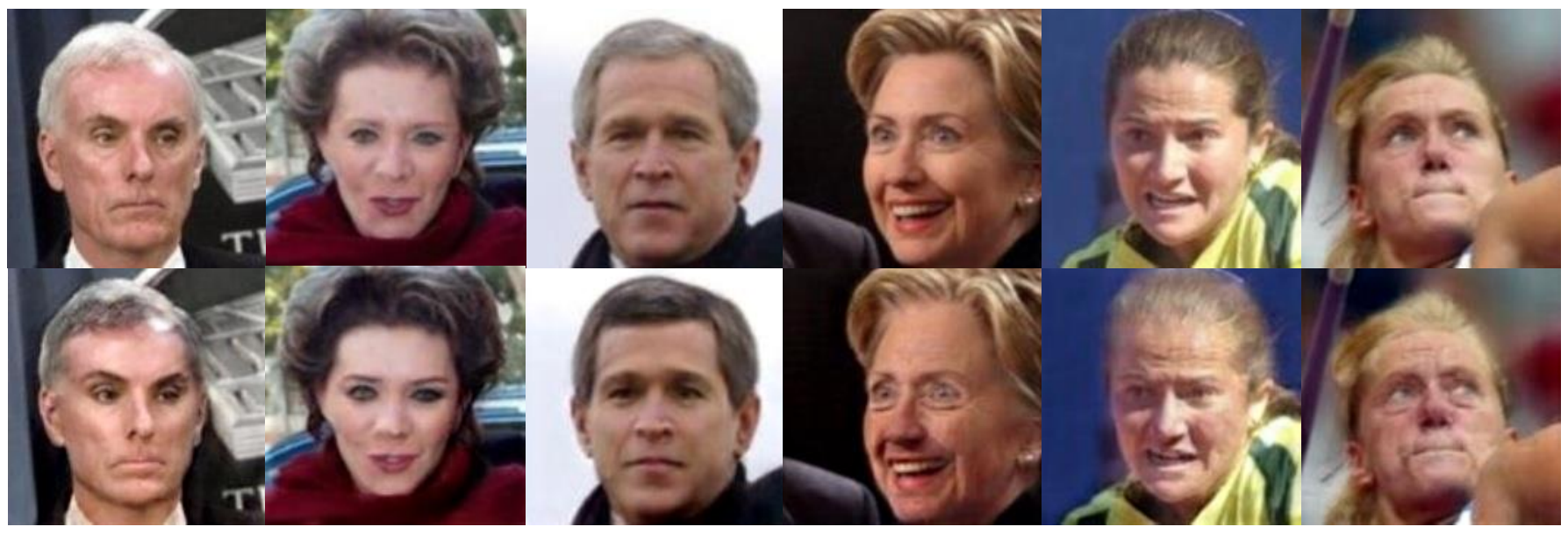}
\subcaption{\textit{Young}}
\end{minipage}
\end{center}
   \caption{Face attribute manipulation on the LFW dataset. For each sub-figure, the first row shows the input images and the second row presents the manipulate images. Results of the primal/dual task of each attribute manipulation are presented in the first/last three columns.}
\label{fig:res_lfw}
\end{figure*}

\section{Experiments}
\subsection{Local and Global Attribute Manipulation}
Among the six attributes, we group \textit{glasses}, \textit{mouth\_open}, \textit{smile} and \textit{no\_beard} as local attributes since the manipulations only operate on local face area. The other two attributes \textit{male} and \textit{young} are treated as global attributes. 
We compare the results of our method and those of the state-of-the-art method VAE-GAN~\cite{larsen2015autoencoding} on the CelebA dataset (Fig.~\ref{fig:res}). The results on the LFW dataset are presented in Fig.~\ref{fig:res_lfw}. 

We firstly give an overall observation of the results. As shown in Fig.~\ref{fig:res}, the VAE-GAN model~\cite{larsen2015autoencoding} changes many details, such as hair style, skin color and background objects. In contrast, the results from our method remain most of the details. Comparing the original images in the first row and the transformed images in the third row, we can find that details in original face images mostly remain the same in their manipulated counterparts except the areas corresponding to the target attributes. This observation is also proved by the residual images in the last row. For local attribute manipulation, the strong responses on the residual images mainly concentrate in local areas. For example, when adding sun glasses to the face image, the most strong response on the residual image is the black sun glasses. Similarly, removing glasses will cause the residual image to enhance the eyes and remove any hint of glasses that is presented in the original face image. 

Local face attribute manipulations are straightforward and obvious to notice. Further, we investigate more interesting tasks such as \textit{mouth\_open} and \textit{smile} manipulations. Both manipulations will cause the \lq\lq movement\rq\rq ~of the chin. From Fig.~\ref{fig:res}(c,d), we can observe that the \lq\lq movement\rq\rq ~of the chin is captured by the image transformation networks. When performing the \textit{mouth\_open} manipulation, the network \lq\lq lowers\rq\rq ~the chin and when performing \textit{mouth\_close} manipulation, the network \lq\lq lifts\rq\rq ~the chin.

The most challenging task would be manipulating global attributes \textit{young} and \textit{male}. The networks have to learn subtle changes such as wrinkles, hair color, beard \etc. In Fig.~\ref{fig:res}(f), changing from young to old will cause more wrinkles and the dual operation will darken the hair color. From the residual images in Fig.~\ref{fig:res}(e), we observe that the main difference between the male and the female are the beard, the color of the lips and the eyes. The strong responses in the residual images for these two manipulations are scattered over the entire images rather than restricted within a local area.


\subsection{Ablation Study}
\begin{figure}
\begin{center}
\includegraphics[width=1.0\linewidth]{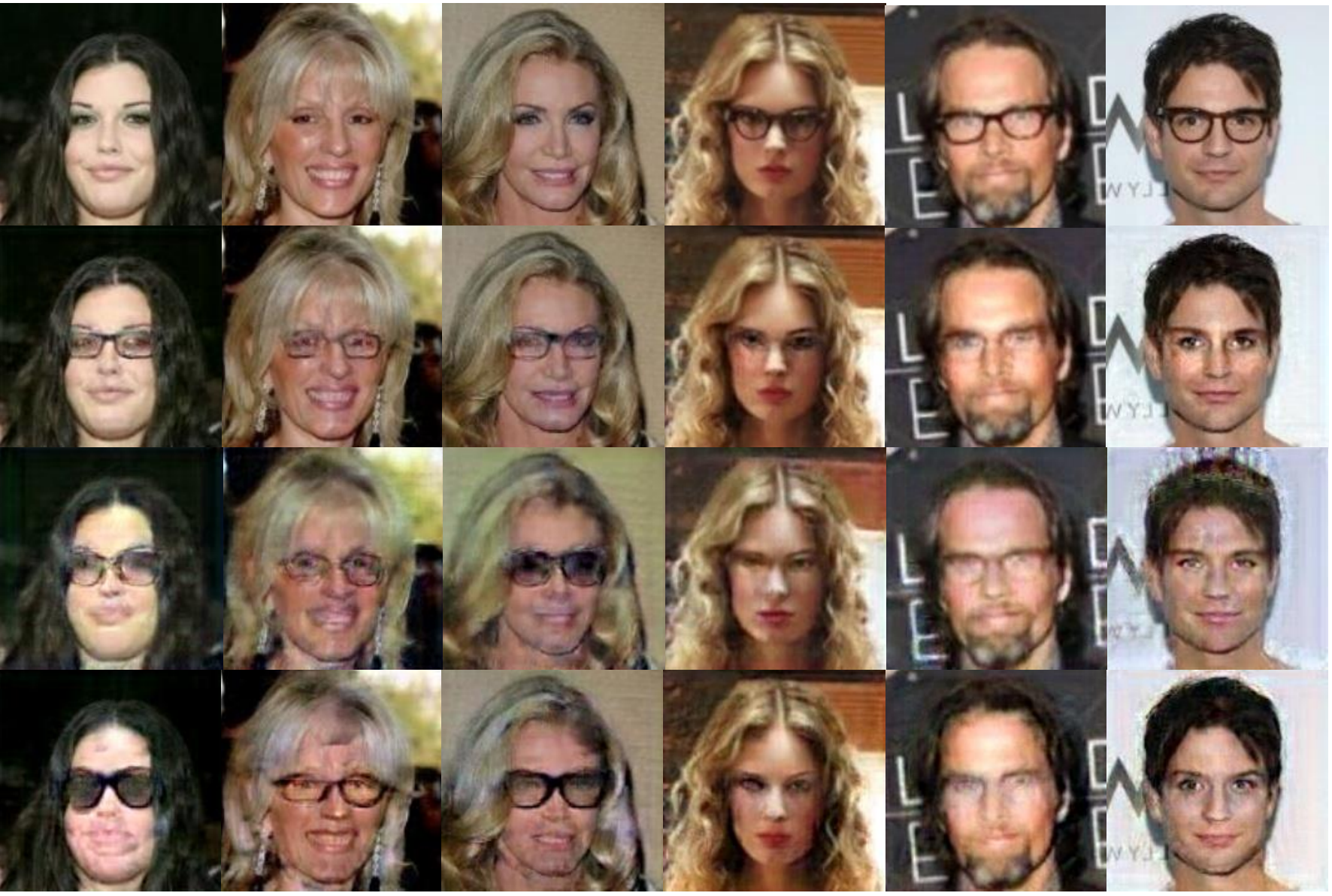}
\end{center}
\caption{Validation of residual image learning and dual learning in \textit{glasses} manipulation. First row: the original input images. Second row: the result images from the proposed model. Third row: the result images from the model without residual image learning. Last row: the result images from the model without dual learning.}
\label{fig:noresdual}
\end{figure}

Our model consists of two pivotal components: residual image learning and dual learning. 
In this section, we further validate their effectiveness. We modify the proposed model to obtain two more models. One breaks the identity mapping in the transformation networks to enforce the networks to learn to generate the entire image. The other breaks the data-feed loop (\ie the output of $G_0$ and $G_1$ will not be fed to each other). Other network settings are kept the same as those of the proposed model. We use the manipulation of \textit{glasses} as an example and the results are shown in Fig.~\ref{fig:noresdual}. We observe that without residual image learning, the model produces much lower quality images in terms of introducing much noise and some correlated features (\eg wrongly added beard in the second and third column), which indicates the task has become challenging. The drop of dual learning also deteriorates image quality. We notice some change in hair color which is caused by the performance degradation of the transformation networks. The effectiveness of the dual learning can be explained from two aspects. 1) Images generated from both generators increase the number of training samples. 2) During the dual learning phase, the ground truth images for $G_1(G_0(x_0))$ and $G_0(G_1(x_1))$ are known, which eases the training of both generators. Thus, we argue that combining residual image learning with dual learning will lead to better  manipulation results.

\begin{figure}
\begin{center}
\includegraphics[width=1.0\linewidth]{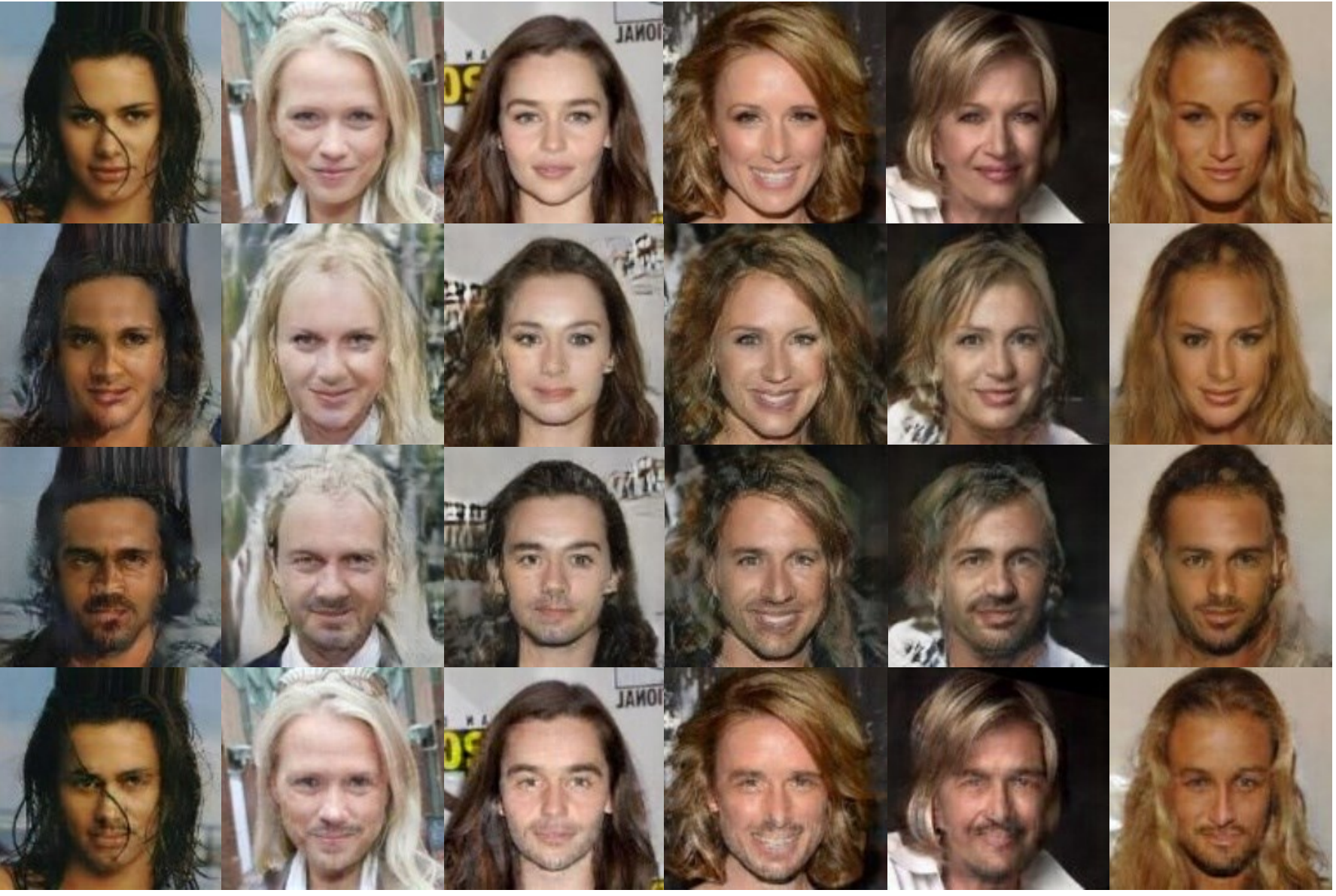}
\end{center}
\caption{Visual feature decorrelation in the manipulation of attribute \textit{no\_beard}. First row: the original input images. Second row: the reconstructed images from the VAE-GAN model~\cite{larsen2015autoencoding}. Third row: the manipulated images from the VAE-GAN model~\cite{larsen2015autoencoding}. Last row: the manipulated images from the proposed model.}
\label{fig:exp_visual_cor}
\end{figure}
\subsection{Visual Feature Decorrelation}
Training a classifier in an end-to-end way does not ensure the classifier can precisely identify the target visual features especially when the dataset is highly biased. Blind spots of predictive models are observed in~\cite{lakkaraju2016discovering}. For example, if the training data consist only of black dog images, and white cat images. A predictive model trained on these data will incorrectly label a white dog as a cat
with high confidence in the test phase. When analyzing the CelebA dataset, we find that \textit{male} and \textit{no\_beard} is highly correlated (the Pearson correlation is -0.5222). It is not surprising since only  male faces have beard.

Classifiers trained with correlated features may also propagate the blind spots back to the generative models which may cause the generators to produce correlated visual features.
To demonstrate that the proposed method can learn less correlated features, we choose to add beard to female face images. Since there are no images containing female faces wearing beard in the training data, this manipulation may introduce other correlated male features. 
We compare the manipulation results from our method with those from the VAE-GAN model~\cite{larsen2015autoencoding} in Fig.~\ref{fig:exp_visual_cor}. 
We show the VAE-GAN reconstruction results of the original images to ensure that the VAE-GAN model do learn well about the original images. However, the hair length in the manipulated images is significantly shorter than that in the original images. This could be explained by the fact that most male faces wear short hair and this visual feature is correlated with the beard feature. The VAE-GAN model incorrectly treats the short-hair feature as the evidence of the presence of the beard. However, the proposed method successfully decorrelates these two features. The hair in the transformed images is almost the same as that in the original images. We owe this appealing property to residual image learning and dual learning that help the method concentrate on attribute-specific area.

\subsection{Landmark Detection with Glasses Removal}
\label{sec:landmark}
Besides visually inspecting the manipulation results, we quantify the effectiveness of glasses removal by the performance gain of face landmark detection accuracy. The landmark detection algorithm is the ERT method~\cite{kazemi2014one} implemented with \textit{Dlib}~\cite{dlib09}. We trained the detection model on the 300-W dataset~\cite{sagonas2013semi}. Three test sets are included into consideration. They are dataset $D1$ containing images wearing glasses, dataset $D0$ containing images without glasses and dataset $D1_m$ containing the same images as $D1$ while those images are processed with glasses removal using the proposed method. Note that $D0$ and $D1$ are the same test sets as those in the experiment of \textit{glasses} manipulation. 

\begin{figure}
\begin{center}
\includegraphics[width=1.0\linewidth]{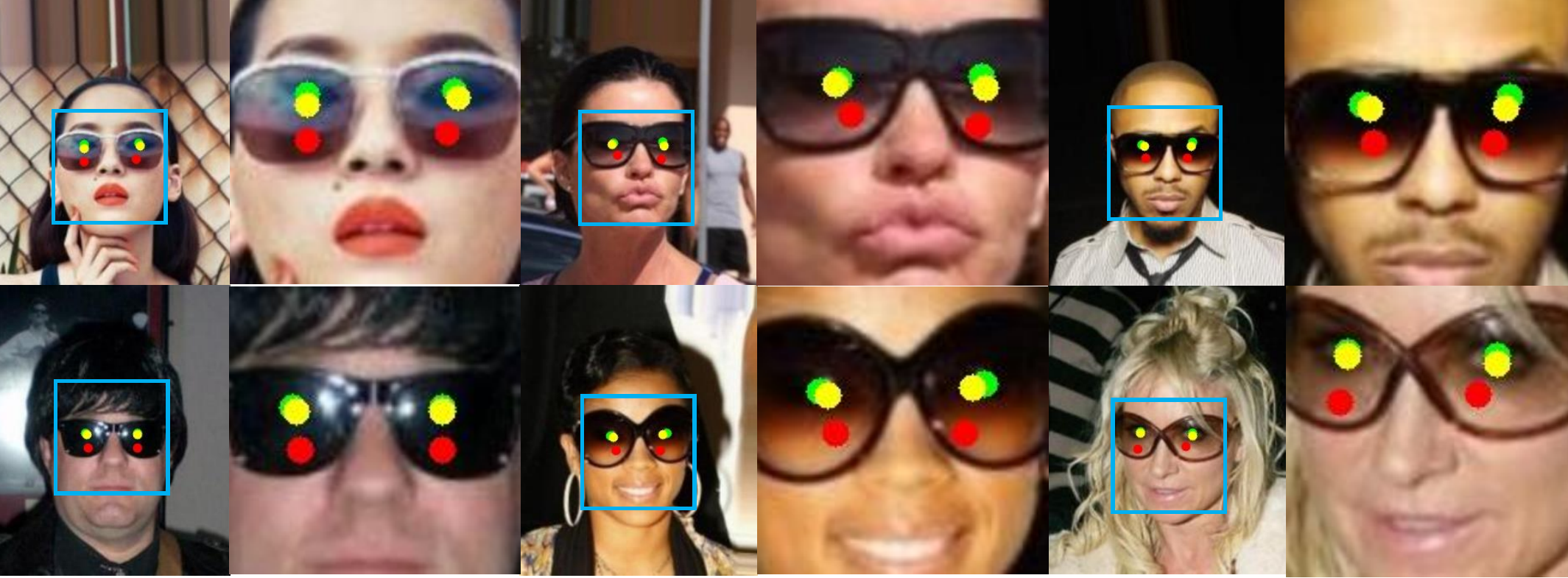}
\end{center}
\caption{Performance gain on landmark detection brought by glasses removal. The ground truth landmarks are shown as green points. The detected landmarks before and after glasses removal are shown as red and yellow points respectively.}
\label{fig:landmark}
\end{figure}

The detection results are reported in Tab.~\ref{tb:exp:landmark} and illustrated in Fig.~\ref{fig:landmark}. Although the identities in $D0$ and $D1$ are different, we can still find that wearing glasses affects the landmark detection. Comparing the results of the eye landmark detections between the first and the second column, we find the detection error increases. However, the error on $D1_m$ is much lower than that on $D1$ which demonstrates the benefit of applying glasses removal for eye landmark detection. Comparing the results in the second row in Tab.~\ref{tb:exp:landmark}, we observe that the errors on $D1$ and $D1_m$ are almost the same which indicates that the rest parts of the face remain almost unchanged. 

\begin{table}
\begin{center}
\begin{tabular}{|c|c|c|c|}
\hline
Landmark & $D0$ & $D1$ & $D1_m$ \\
\hline
\hline
Eye landmarks & 0.02341 & 0.03570 & 0.03048 \\
\hline
Rest landmarks & 0.04424 & 0.04605 & 0.04608 \\
\hline
\end{tabular}
\end{center}
\caption{The average normalized distance error from the landmark detection algorithm on $D0$, $D1$, and $D1_m$. \lq\lq Eye landmarks\rq\rq ~means the landmarks of the left and right eyes. \lq\lq Rest landmarks\rq\rq ~indicates the landmarks of the nose and the left and right corners of the mouth. }
\label{tb:exp:landmark}
\end{table}

\section{Discussion}
\begin{figure}
\begin{center}
\includegraphics[width=1.0\linewidth]{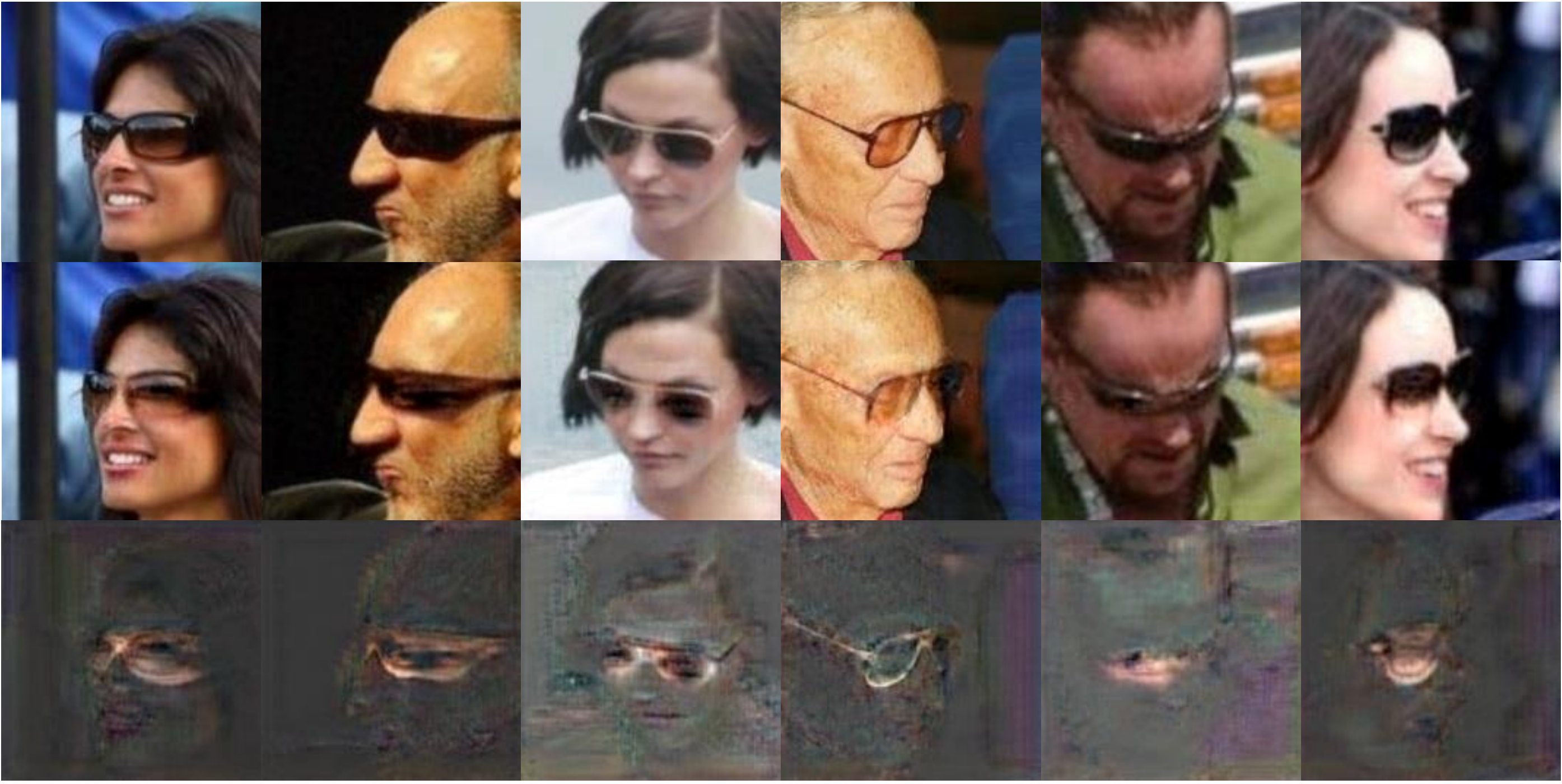}
\end{center}
\caption{Glasses removal failure cases due to large poses. First row: the original input images. Second row: the manipulated images. Third row: the residual images.}
\label{fig:failure_cases}
\end{figure}

The hyper-parameters $\alpha$ and $\beta$ are manually set in this work. 
We empirically found $\alpha$=5e-6 and $\alpha$=5e-4 are appropriate for global and local attribute manipulation respectively. Larger $\alpha$ and $\beta$ will prevent the generative models from simulating the manipulations and will collapse the models to 0. The adoption of only one discriminator in our work is mostly for improving the computation efficiency. There could be identity overlaps between the training set and the test set since the number of identities is much smaller than the number of all images and images for testing are also randomly selected. However, even the face identities could be the same, the face poses, illumination conditions, and the background objects can hardly be the same. Therefore, there is little information leak caused by identity overlaps to the test set. In Fig.~\ref{fig:failure_cases}, we show some failure cases in glasses removal. Those cases are mainly due to large poses.

\section{Conclusion}
In this work, we present a GAN based method to tackle the task of face attribute manipulation. We adopt both residual image learning and dual learning in our framework. Those strategies allow image transformation networks to focus on attribute-specific area and learn from each other. Experiments demonstrate that the proposed method can successfully manipulate face images while remain most details in attribute-irrelevant areas unchanged. 

{ \small
\bibliographystyle{ieee}
\bibliography{egbib}
}

\end{document}